\documentclass[fleqn,10pt]{wlscirep}
\usepackage[utf8]{inputenc}
\usepackage[T1]{fontenc}
\usepackage{graphicx}
\usepackage{subcaption}
\usepackage{placeins}
\usepackage{cite}
\usepackage{verbatim}

\title{Pattern-Based Phase-Separation of Tracer and Dispersed Phase Particles in Two-Phase Defocusing Particle Tracking Velocimetry}

\author[1,2]{Christian Sax}
\author[1]{Jochen Kriegseis}
\affil[1]{Institute of Fluids Mechanics (ISTM), Karlsruhe Institute of Technology (KIT), Kaiserstraße 10, 76131 Karlsruhe, Germany}
\affil[2]{Karlsruhe Institute of Technology, Institute of Applied and Numerical Mathematics, Karlsruhe, 76133, Germany}
\affil[*]{jochen.kriegseis@kit.edu}

\begin{abstract}

This work investigates the feasibility of a post-processing-based approach for phase separation in defocusing particle tracking velocimetry for dispersed two-phase flows. The method enables the simultaneous 3D localization determination of both tracer particles and particles of the dispersed phase, using a single-camera setup. The distinction between phases is based on pattern differences in defocused particle images, which arise from distinct light scattering behaviors of tracer particles and bubbles or droplets. Convolutional neural networks, including Faster R-CNN and YOLOv4 variants, are trained to detect and classify particle images based on these pattern features. To generate large, labeled training datasets, a generative adversarial network based framework is introduced, allowing the generation of auto-labeled data that more closely reflects experiment-specific visual appearance. Evaluation across six datasets, comprising synthetic two-phase and real single- and two-phase flows, demonstrates high detection precision and classification accuracy (95–100\%), even under domain shifts. The results confirm the viability of using CNNs for robust phase separation in disperse two-phase DPTV, particularly in scenarios where traditional wavelength-, size-, or ensemble correlation-based methods are impractical.

\end{abstract}
\begin{document}

\flushbottom
\maketitle
\thispagestyle{empty}

%\chapter{Simultaneous Measurement of two Phases in Defocusing Particle Tracking Velocimetry}
%\label{chap:NN_2Phase}

\section{Introduction}

%why care about both phase particles? (hy problem matters)
In dispersed two-phase flows (DTPFs), the measurement of the velocity of both the dispersed and continuous phases is essential. Techniques such as particle image velocimetry (PIV) and particle tracking velocimetry (PTV) are widely employed to obtain flow fields from tracer particles. In two-phase systems, the motion of both phases is often of interest in order to more accurately characterise the underlying phenomena.
%examples why having both phases is important
Accurate measurement of the slip velocity, i.e.\ the velocity difference between the dispersed phase particles and the carrier phase, constitutes critical information for the modeling of DTPFs, particularly under non-uniform flow conditions \cite{Balachandar_2024,Brennen2005,Minier2025}.
The velocity difference between the phases influences, for example, the system behavior of the flow at the outlet of nozzles \cite{Balachandar_2024}, the mixing behavior and heat transfer in bubble column reactors \cite{JOSHI2001}, the overall dynamics of bubble columns \cite{GRIENBERGER1992}, phase change in screw compressors \cite{Leister_2022,LangeSax2024b_conf}, dispersion and momentum exchange in sprays \cite{GOUNDER2012}, and the clustering and evaporation of droplets \cite{Hardalupas2018}.

%what work has been done and how? (what people have tried)
Several methods have been developed to measure both phases simultaneously, often by combining different techniques. The present work focuses on methods akin to PIV and PTV, which enable non-intrusive field measurements. PTV provides Lagrangian data by tracking individual particles, whereas PIV yields Eulerian velocity fields. Their application to two-phase flows has been extensively studied; see, for example, \cite{Cerqueira2018,Kitagawa2007,Kiger2000,CHEN1992,Jakobsen1996}. 
A central concept in two-phase PIV or PTV is the presence of two distinct particle types: tracer particles that follow the continuous phase, and particles of the dispersed phase, such as bubbles, droplets, or solids. The reliable distinction between these particle types is essential for phase-specific measurements.
In essence, two types of approaches are employed to distinguish between the particle types:

%wavelength based
The first approach employs wavelength-based techniques to distinguish between particle types. Dual-camera setups utilize wavelength filters to differentiate between particles with or without fluorescence or with fluorescence at different wavelengths \cite{Cerqueira2018,Kosiwczuk2005}. Alternative methods exploit the separate color channels of a single RGB camera for phase distinction \cite{Kitagawa2007}. Other techniques apply laser-induced fluorescence (LIF) to identify, for example, bubbles by detecting the absence of tracers in the carrier fluid.
Wavelength-based techniques are advantageous due to their reliable phase distinction. However, multi-camera systems are complex and challenging to calibrate, and RGB cameras offer only lower sensor resolution.

%postprocessing based approaches
The second category of approaches distinguishes the phases solely during data post-processing. These methods include exploiting size differences between particle types \cite{Kiger2000,CHEN1992,Jakobsen1996,Khalitov2002}, combining differences in particle size and the intensity of scattered light \cite{Khalitov2002}, and applying ensemble correlation of particle motion \cite{DELNOIJ2000,Seol2008} to separate the phases.
Post-processing-based distinction generally permits the use of standard equipment, but depends on the classification accuracy of the underlying method. Ensemble-based techniques require a sufficient number of particles within the image, and size-based distinction is only feasible when the two particle types differ significantly in size.

%bring the defocusing PTV here
Since DTPFs are inherently three-dimensional, their investigation should ideally be based on volumetric measurement techniques. While conventional two-phase PIV is typically limited to planar measurements, the same phase separation strategies can, in principle, be extended to three-dimensional PTV methods such as tomographic PTV \cite{Nishino1989,Maas1993,Schanz2013,Schanz2016}, enabling full 3D-3C (three dimensional - three components) Lagrangian velocity measurements.
In applications with limited optical access, single-camera techniques such as defocusing PTV (DPTV) \cite{Willert.1992,Olsen2000} and astigmatism PTV (APTV) \cite{Cierpka2010,KAO1994} offer volumetric measurement capabilities. These methods determine the out-of-plane position of particles based on the degree of defocus in the particle image (PI). The present work focuses on the phase distinction in DPTV. However, the phase distinction in DPTV is less trivial:
%Why phase separation not so trivial in DPTV
Although phase separation by wavelength is feasible in APTV and DPTV, it undermines the principal advantage of these methods being single-camera systems. Moreover, the use of RGB cameras can be problematic due to their lower sensor resolution, which reduces the accuracy of PI diameter estimation. 
Post-processing-based phase separation presents an alternative, however, distinguishing particles by size is particularly challenging in defocused images, as PI size is primarily governed by the degree of defocus rather than the actual particle size. Ensemble-based methods are likewise unsuitable for DPTV, as they require high seeding densities, which are often unfeasible due to the method’s inherent limitation to lower seeding levels caused by PI overlap.

%Our approach the basic idea.
The present work proposes a post-processing approach for two-phase DPTV that employs neural networks to classify PIs based on their visual patterns. The method enables 3D position determination of both tracers and bubbles or droplets, using a single camera by exploiting differences in the appearance of defocused PIs, which result from distinct scattering behaviors.
By relying on scattering characteristics rather than particle size, the approach permits tracer and dispersed phase particles to be of identical size and eliminates the need for specialized equipment such as RGB cameras, which may impose other limitations. As with any post-processing method, the effectiveness of the approach depends on the classification accuracy and the algorithm’s ability to generalize across different particle types.

%why the CNNs
For the task of pattern-based phase distinction, the present approach employs convolutional neural networks (CNNs) \cite{Lecun.1990,Lecun.1998}, which are increasingly adopted within the DPTV and APTV communities and have demonstrated enhanced performance in PI detection \cite{Cierpka.2019,Franchini.2020,Konig.2020,Barnkob.2021,Dreisbach.2022,Sax2023c,Zhang2023,Ratz2023}. Modern object detection networks, such as Faster R-CNN \cite{Ren.2017} and YOLO \cite{Redmon2016,Redmon2017,Redmon2018}, are capable not only of detecting but also classifying objects, rendering them a natural choice for distinguishing between different PI types.
This work utilities object detection networks to detect and classify different PI types based on their visual patterns. While this distinction is theoretically applicable to various particle types, the present study focuses exclusively on differentiating between tracer particles and droplets or bubbles. Consequently, the use cases addressed pertain specifically to gas-liquid DTPFs.
 
%why bother about CNN training data? --> generalization capability and need for retraining
Nevertheless, the application of CNNs for pattern-based phase distinction in this context introduces several non-trivial challenges.
While CNNs are effective for PI detection in DPTV, their performance is typically highest when the test data closely resembles the training data. However, variations in optical setups can significantly alter the appearance of PIs, thereby challenging the network’s ability to generalize across different conditions and potentially reducing both detection and classification accuracy.
%idea to build diverse training set and method to autolabel training data.
To enhance generalization, the training dataset should be large and diverse, incorporating images from a variety of optical configurations. However, object detection networks typically require labeled data to be trained. The acquisition of such labeled data remains one of the principal barriers to the broader adoption of CNNs in DPTV. 
Alternatively, training data tailored to a specific experiment can be used to train a CNN on case-specific data, thereby reducing the need for generalization. However, this approach necessitates a method for acquiring and labeling experiment-specific training data.
To address the problem of acquiring labeled training data for CNNs in DPTV, the present work proposes a framework for generating large volumes of auto-labeled training data using generative adversarial networks (GANs) \cite{Goodfellow2014}. To evaluate the feasibility of CNN-based phase separation in dispersed two-phase DPTV, Faster R-CNN and YOLOv4 \cite{Bochkovskiy2020} are employed. These networks are assessed based on their ability to detect and classify PIs under both familiar and previously unseen imaging conditions.

%perspective 
In summary, the present study introduces a method for the pattern-based distinction of tracers from bubbles or droplets using CNNs in two-phase DPTV, while concurrently addressing the principal challenge of acquiring suitable training data for these networks.
This approach may also be extended by integrating interferometric particle imaging (IPI) \cite{Konig.1986,Glover.95,Niwa.2000}, enabling simultaneous determination of the particle size of the dispersed phase from fringe patterns in the PIs.

\section{Theoretical Concept of Pattern based Phase Distinction in Defocused Images}

\begin{figure} [htb]
\centering
    \includegraphics[width=0.9\textwidth]{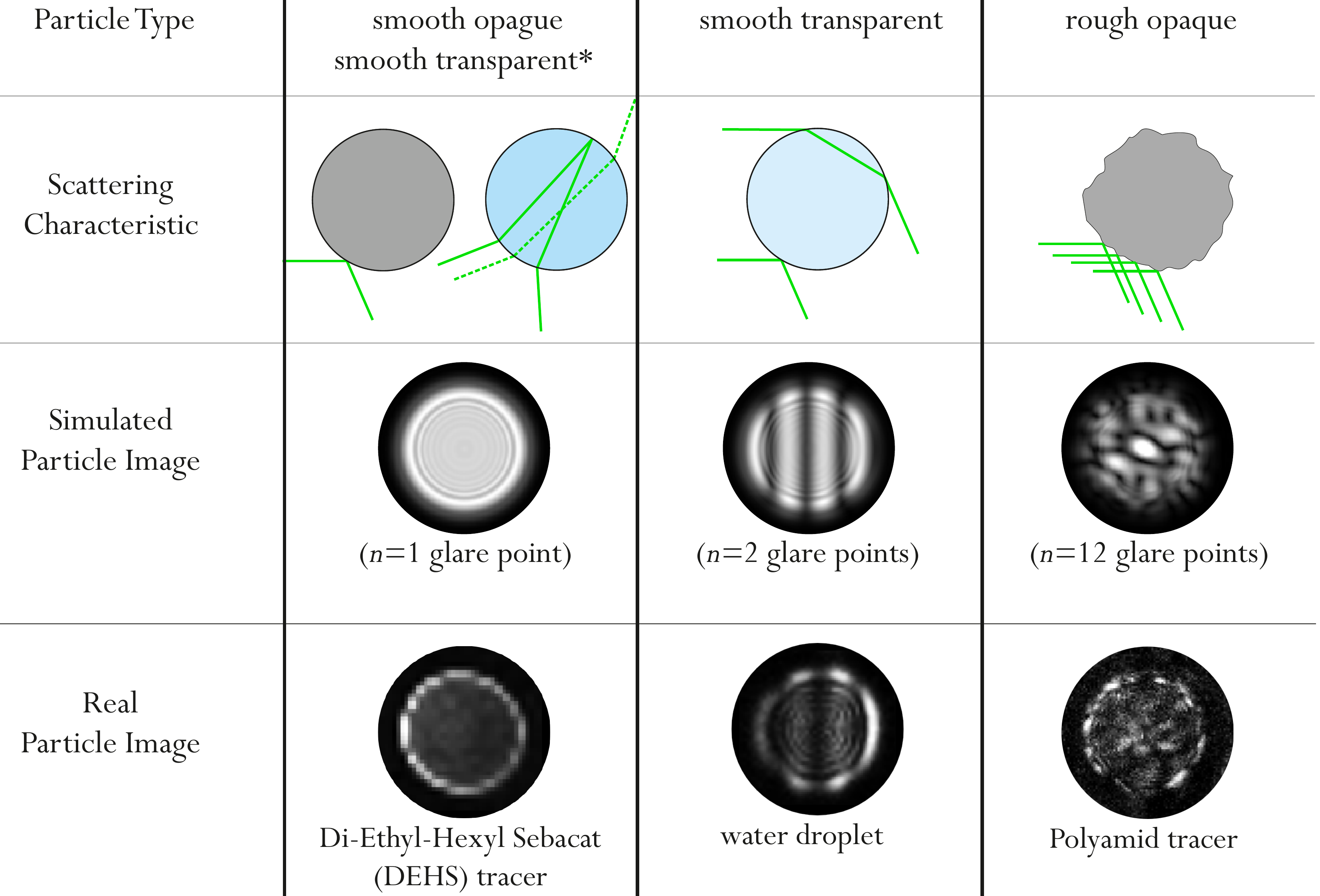}
\caption{Three different types of PIs: PIs with an empty pattern (left), PIs with a regular unidirectional pattern (middle) and PIs with a irregular speckle pattern (right). The particle type, the scattering characteristic and simulated and real examples are shown. The simulated examples where created using the forward model from Sax\,\textit{et al.}\,\cite{Sax2025_IP} and the glare point based scattering characteristic. For the rough opaque particle a number of glare points significantly larger than two (here 12) where chosen to model multiple randomly distributed glare points. The real examples show a DEHS tracer from Exp.\,G\ref{exp:G2} (left), a water droplet from Exp.\,T\ref{exp:T4} (middle) and a Polyamid tracer from Exp.\,G\ref{exp:G1} (right).}
\label{fig:Chap5_PI_Theory_Real}
\end{figure}

%idea 1 vs. two vs. multiple irregular glare points 
The present two-phase approach for DPTV utilities the pattern of a PI to differentiate between tracers and bubbles or droplets. While defocused PIs of tracers and those of bubbles or droplets are fundamentally similar, the key distinction lies in the presence of unidirectional interference fringes in the latter, as typically observed in IPI. In contrast, PIs from opaque tracer particles, as used in DPTV, generally do not exhibit such fringe patterns. It should be noted, however, that certain transparent tracers may also produce fringe patterns, this issue is addressed later in the present work.
These differing PI patterns arise from the distinct scattering characteristics of the particles. These characteristics, in turn, result from differences in particle surface properties and transparency, as illustrated in Fig.\,\ref{fig:Chap5_PI_Theory_Real}. Fig.\,\ref{fig:Chap5_PI_Theory_Real}  illustrates the three types of PI patterns along with their respective scattering characteristics. 

For example, an opaque tracer may possess either a smooth or a rough surface, each influencing the scattering behavior differently.
%smooth opaque tracer 
In the case of a smooth opaque tracer, light is reflected towards the camera at only a single point (glare point), resulting in a PI that exhibits no interference pattern, referred to as an \textit{empty pattern}; see the left column of Fig.\,\ref{fig:Chap5_PI_Theory_Real}. Similarly, smooth, transparent tracers also exhibit only a single glare point under specific illumination angles. An example of such tracers is Di-Ethyl-Hexyl Sebacat (DEHS), which typically produces empty pattern PIs.

%rough surface
For tracers with a rough surface, light is scattered at multiple points across the surface, producing glare points without a geometrical order. These rough particles exhibit a speckle pattern \cite{Dainty1975}, which appears as a random, non-directional pattern (\textit{irregular pattern}); see the right column of Fig.\,\ref{fig:Chap5_PI_Theory_Real}. Polyamid tracers are an example of particles that produce such speckle patterns.

%droplet or bubbles
In contrast, bubbles or droplets, which are transparent particles with smooth interfaces, reflect and refract light at multiple points. This results in more than one glare point on the particle surface \cite{vandeHulst.1991}. Notably, these glare points lie along a single line defined by the scattering plane, producing unidirectional interference fringes in the PI (\textit{regular pattern}); see the middle column of Fig.\,\ref{fig:Chap5_PI_Theory_Real}. Thus, PIs of bubbles or droplets can be identified by their regular, unidirectional fringe patterns.

It should be noted that certain transparent tracers (e.g. DEHS tracers) may also possess smooth surfaces and can therefore exhibit unidirectional interference patterns, making them difficult to distinguish from bubbles or droplets. In such cases, alternative tracer materials should be selected. Occasionally, a smooth transparent tracer with a refractive index different from that of the bubbles or droplets may still produce an empty pattern PI. This occurs under specific scattering angles where only a single glare point is formed, while other particles of different refractive index may exhibit two. However, this scenario is complex and requires careful consideration of the scattering characteristics of transparent particles.
 
%which kind of flows can be characterized 
Within the scope of this work, only gas-liquid DTPFs, such as bubbly flow, are considered. Consequently, neural networks were trained to distinguish between bubbles or droplets, characterized by regular, unidirectional fringe patterns, and tracers, which exhibit either empty or irregular speckle patterns. Accordingly, PIs with empty and speckle patterns were grouped into a single class (tracers), while PIs from bubbles and droplets were assigned to the other class (dispersed phase). The approach may also be applicable to other flow types, however, these lie beyond the scope of the present work and remain a subject for future investigation.

\section{Auto-labeling Approach and Training of Networks for Particle Detection and Classification}

To employ CNNs for the detection and classification of different types of PIs, training data must first be acquired and labeled. In the context of DPTV, at least three distinct approaches have been reported in the literature for obtaining such training data:
%real data
Images can be taken from calibration measurements to train the CNN \cite{Cierpka.2019}. However, labeling experimental data is time-consuming, since there is no known ground truth and typically results in suboptimal dataset sizes.
%synthetic data
An alternative is training on synthetic (model based) data \cite{Franchini.2020,Sax2023c}. Using physical models to generate synthetic images with known ground truth allows for automatic labeling and the creation of sufficiently large training datasets. However, synthetic images do not necessarily represent real images with sufficient fidelity and often lack diversity in features, which may result in poor generalization capability of the network. Additionally, large dataset sizes are only partially beneficial if the variety in features is limited and the images become repetitive.
%deep fakes
A compromise between the two approaches is the use of image-generating deep learning methods, such as GANs, to generate synthetic (data based) images that more closely resemble real PIs \cite{Dreisbach.2022}.

\begin{figure}[htb]
    \centering
    \includegraphics[width=\linewidth]{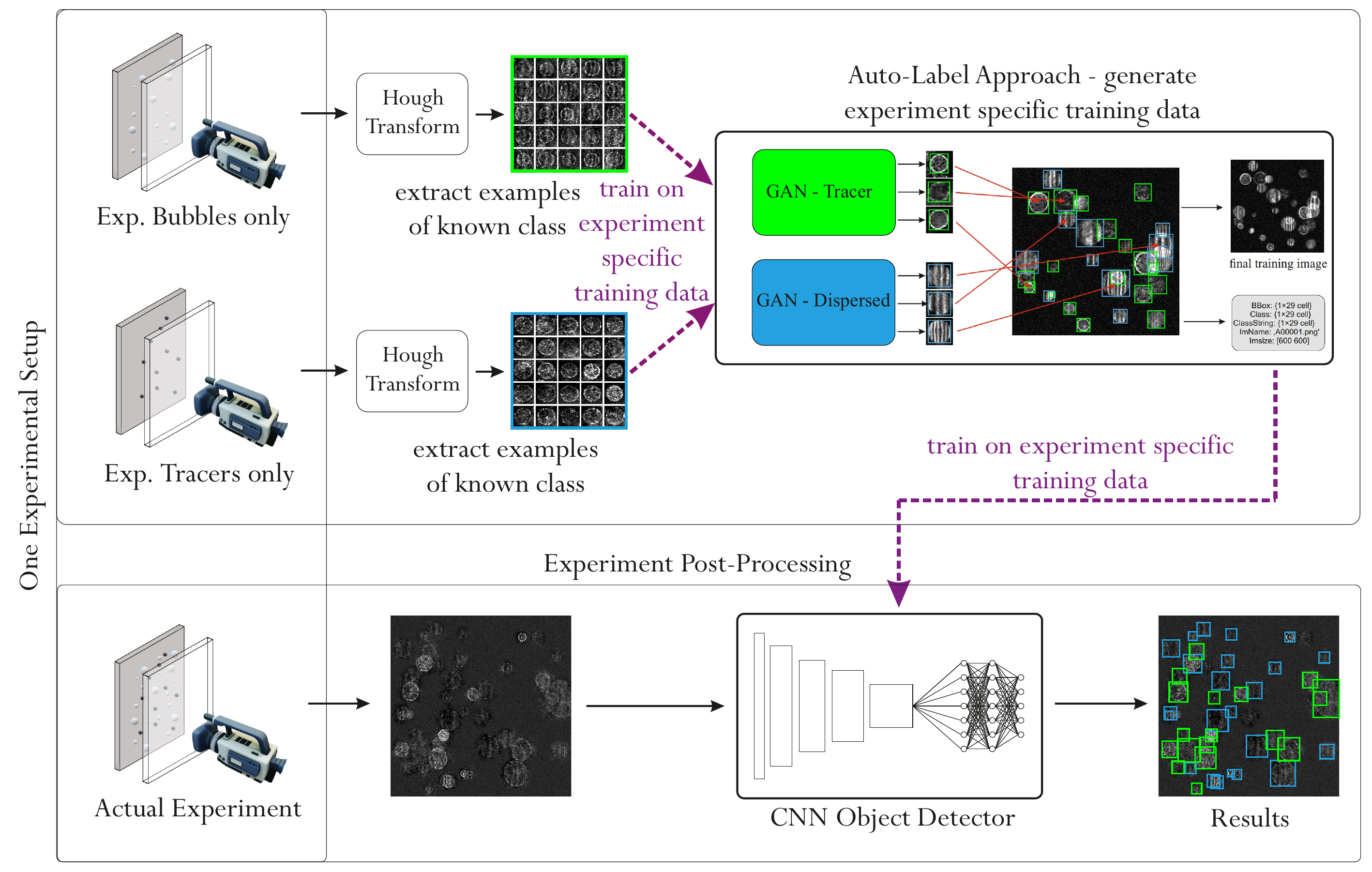}
    \caption{Auto-labeling approach and PI detection. Two data acquisition experiments are conducted, each involving a single phase. PIs are extracted from the resulting image sets using conventional algorithms. These extracted PIs are then used to train GANs, to generate mixed-phase image sets. The mixed-phase sets are subsequently used to train an object detection CNN. Finally, the trained CNN is applied to the actual two-phase experimental data.}
    \label{fig:Chap5_AutoLabelApplication}
\end{figure}

%our approach
An approach based on GAN-generated training data is proposed for object detection in two-phase flow experiments. An overview of the methodology is provided in Fig.~\ref{fig:Chap5_AutoLabelApplication}. The objective is to train a CNN on data that closely replicates the conditions of the actual experiment.
First, two data collection experiments are conducted using the same experimental setup as the target experiment (right side of Fig.~\ref{fig:Chap5_AutoLabelApplication}). Each experiment captures only a single phase, either tracers or the dispersed phase. PIs are extracted from the resulting image sets using conventional algorithms. These extracted PIs are then used to train two GANs, each corresponding to a specific PI type (e.g., tracer or bubble). These GANs are then employed to generate synthetic images containing both PI types (upper part of  Fig.~\ref{fig:Chap5_AutoLabelApplication}).
The synthetic mixed-phase images are used to train the object detection CNN. Once trained, the CNN is applied to the actual two-phase experimental data to detect and distinguish between different PI types (lower part of  Fig.~\ref{fig:Chap5_AutoLabelApplication}).

This approach enables the generation of training data tailored to a specific experimental setup, as illustrated in Fig.~\ref{fig:Chap5_AutoLabelApplication}. Alternatively, multiple data collection experiments can be performed to produce more diverse training data, thereby facilitating the development of object detectors capable of generalizing across various experimental configurations.

The approach can be divided into two steps: the first part involves the acquisition of raw images and the training of the GANs for the auto-labeling approach (upper part of Fig.\,\ref{fig:Chap5_AutoLabelApplication}), which is presented in Section\,\ref{chap5:Sec:Approach:Sub:GANTraining}.
The second part concerns the training of the actual object detection network used for phase distinction (lower part of Fig.\,\ref{fig:Chap5_AutoLabelApplication}), which is described in Section\,\ref{chap5:Sec:Approach:Sub:CNNTraining}. 

However, before the training and data acquisition procedures are discussed, the method for generating training data using GANs is elaborated in more detail in Section\,\ref{chap5:Sec:Approach:Sub:Autolabel} (auto-label approach - right upper part of Fig.~\ref{fig:Chap5_AutoLabelApplication}).

\subsection{Conceptual Approach for the Generation of Auto-labeled Training Data}
\label{chap5:Sec:Approach:Sub:Autolabel}

\begin{figure*} [htb]
\centering
    \includegraphics[width=0.9\textwidth]{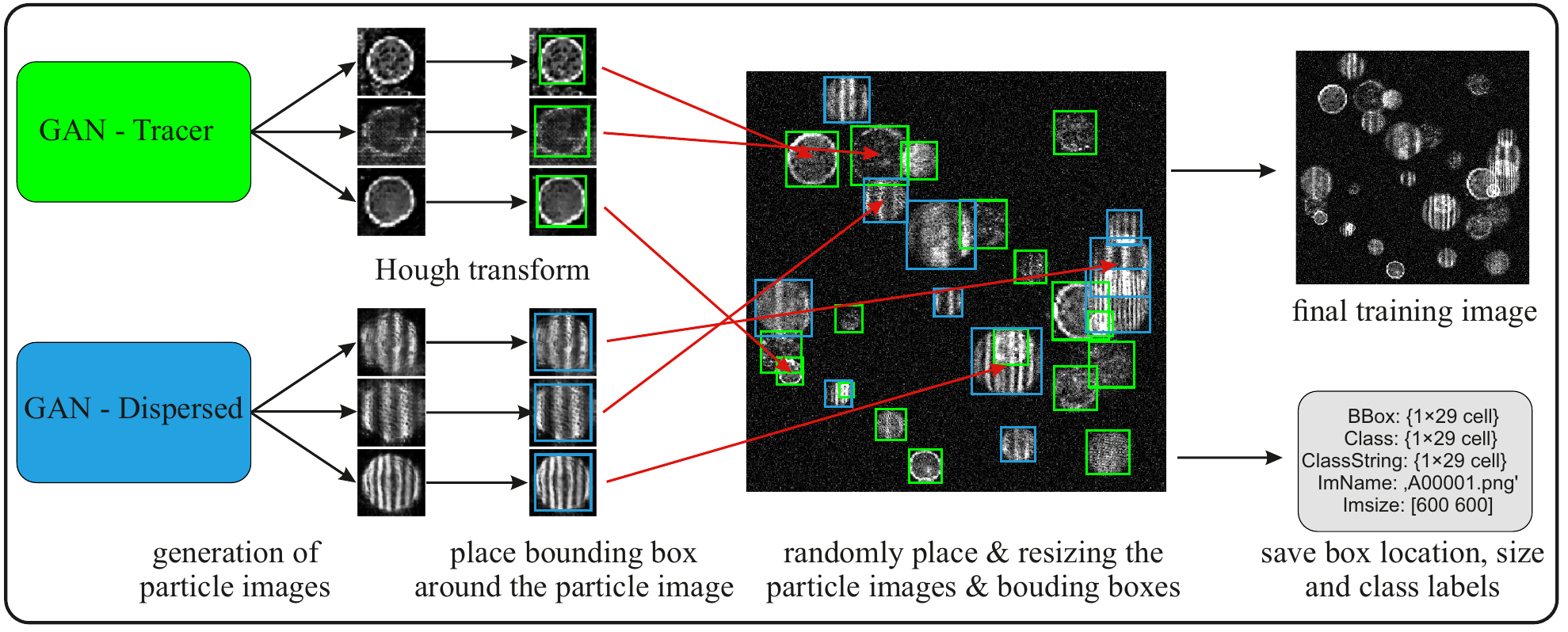}
\caption{Generation of automatically labeled training data for object detection networks. Two GANs - one for tracer particles and one for the dispersed phase - independently generate small image snippets, each containing a single PI. For each snippet, a BB is placed over the PI. The image snippets and corresponding BBs are then randomly resized and inserted into an empty image.}
\label{fig:Chap5_ImageGenScheme}
\end{figure*}

This section provides more detail on the auto-labeling approach using GANs, shown in Fig.\,\ref{fig:Chap5_ImageGenScheme}. The auto-labeling approach employs two GANs trained on PIs of either tracers or bubbles, obtained from dedicated data acquisition experiments. The method operates by using a GAN for each phase to generate a PI, which is then inserted into a full image.
%GAN + Hough
Two GANs, one per PI type, are used so that the class of each generated PI is always known. This also offers the advantage that each GAN only needs to specialize in the features of a single PI type, avoiding the risk of feature mixing between classes.

%get image data for the GANs
To train the GANs, single-phase data acquisition experiments are conducted to collect a large number of images containing only one PI type. These experiments must be single-phase so that the class of the PI in each image is known, and classification can be avoided. A conventional algorithm, such as the circular Hough transform \cite{Atherton.1993,Atherton.1993b,Atherton.1999}, is then used to extract individual PIs from the images, see the Hough transform example in Fig.~\ref{fig:Chap5_AutoLabelApplication}. The Hough transform also determines the size of the PIs, allowing them to be rescaled to a standard size, usable by the GAN.
It should be noted that the data acquisition experiments should be conducted at low seeding densities to ensure reliable PI detection by the Hough transform, minimising false positives and avoiding PI overlap. PI overlap is further mitigated by calculating the intersection over union (IoU) for each detection and removing overlapping PIs from the dataset.
In this way, a large number of individual PI examples can be collected for each PI type. While approach requires the recording of typically 100 to 1000 images per class, it eliminates the need for manual labeling.

%train the GANs
Using these two PI sets, one for tracers and one for bubbles or droplets, the GANs can be trained to generate images containing PIs that resemble those in the recorded datasets. Once trained, the GANs generate image snippets, each containing a single PI of the corresponding type. This is achieved by also providing the GANs only with image snippets of individual PIs in training.
%BB placement
For each GAN-generated image, the Hough transform is applied again to place a bounding box (BB) around the PI. The class label is automatically assigned based on the GAN that generated the image (i.e. tracer-GAN or bubble-GAN); see the BB placement step in Fig.~\ref{fig:Chap5_ImageGenScheme}. 

%pasting into the image
The labeled PIs are then randomly resized using linear interpolation (up and down sampling) and placed at random positions within the image; see the red arrows in Fig.~\ref{fig:Chap5_ImageGenScheme}. The position and size of the PI BB are updated during the placement and resizing process to reflect their location in the full image. In this way, the BB position is known in the final image.

%circular mask 
Before inserting the PI into the image, a circular mask is applied to the GAN-generated image snippet outside the BB to subtract the background. This step reduces artifacts caused by pasting the snippet into the image and prevents the network from learning these artifacts as cues for PI detection. However, there remains a risk that the network may learn the mask itself as an unintended feature.

Using this method with two GANs, a random number of PIs per phase is placed into each image, with every PI assigned a BB and class label. This process can be repeated across multiple images to generate large training datasets that resemble those obtained from the data acquisition experiments.
The advantage of using GANs over directly employing unlabeled PIs from the experiment lies in their ability to produce a greater number of images than the original dataset used for training. In this way, the GANs act as a multiplicative factor, effectively increasing the size of the dataset.

\subsection{Training the Generative Adversarial Networks for Image Generation}
\label{chap5:Sec:Approach:Sub:GANTraining}

To realize this approach, raw images must be collected from data acquisition experiments to train the GAN networks. This section outlines the data acquisition procedures employed in this work, as well as the subsequent training of the GANs.
While the proposed method can be used to tailor training data to a specific experimental setup, the resulting insights on the performance would be highly case-specific.
Instead, a second strategy is adopted: generating a larger and more diverse training dataset by incorporating PIs from multiple experiments. This enables the CNNs to be trained for applicability across different experimental conditions and allows their generalization capability to be evaluated. As a result, the findings presented in this work are more transferable and relevant to a broader range of use cases.

\subsubsection*{Training Data Acquisition}

\begin{figure}[htbp]
    \centering
    \begin{subfigure}[b]{0.23\linewidth}
        \centering
        \includegraphics[width=\linewidth]{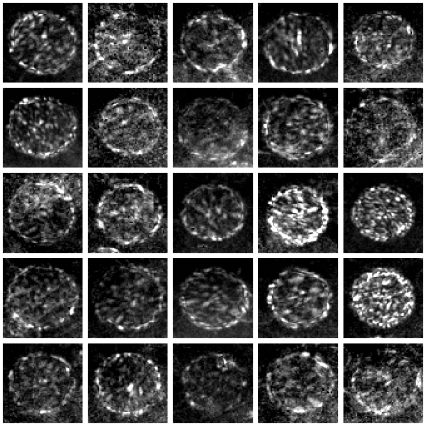}
        \caption{Exp.\,G1: Tracers}
        \label{fig:Chap5_GAN-Train_ExpG1}
    \end{subfigure}
    \hfill
    \begin{subfigure}[b]{0.23\linewidth}
        \centering
        \includegraphics[width=\linewidth]{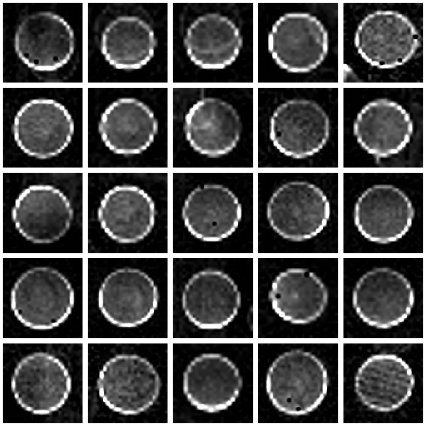}
        \caption{Exp.\,G2: Tracers}
        \label{fig:Chap5_GAN-Train_ExpG2}
    \end{subfigure}
    \hfill
    \begin{subfigure}[b]{0.23\linewidth}
        \centering
        \includegraphics[width=\linewidth]{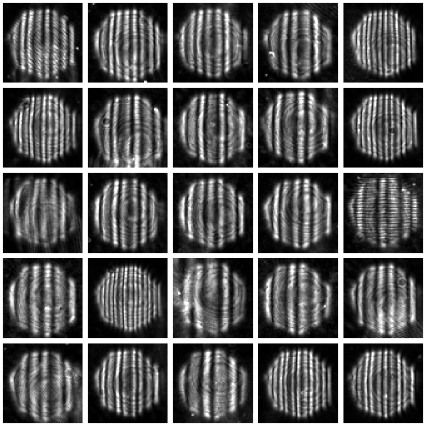}
        \caption{Exp.\,G3: Bubbles}
        \label{fig:Chap5_GAN-Train_ExpG3}
    \end{subfigure}
    \hfill
    \begin{subfigure}[b]{0.23\linewidth}
        \centering
        \includegraphics[width=\linewidth]{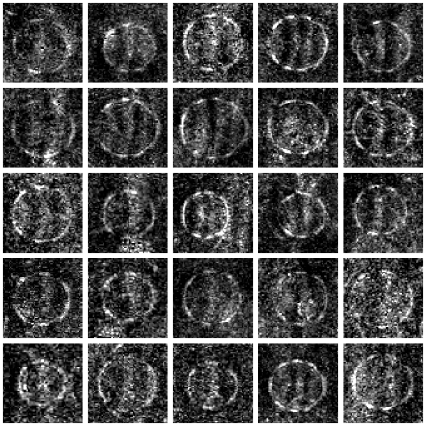}
        \caption{Exp.\,G4: Bubbles}
        \label{fig:Chap5_GAN-Train_ExpG4}
    \end{subfigure}
    \caption{PIs sampled from the Exp.\,G\ref{exp:G1}\cite{LangeSax2024b_conf} (a), Exp.\,G\ref{exp:G2}\cite{Pasch2024} (b), Exp.\,G\ref{exp:G3}\cite{Sax2023d} (c) and Exp.\,G\ref{exp:G4}\cite{LangeSax2024b_conf} (d). These PIs were used to train the GANs for tracer particles (a)\&(b) and for particles of the dispersed phase (c)\&(d).}
    \label{fig:Chap5_GAN-Train_ExpGx}
\end{figure}

In this work, training images were collected from four different experiments: two containing only tracers (Exp.\,G\ref{exp:G1} and G\ref{exp:G2}) and two containing only bubbles (G\ref{exp:G3} and G\ref{exp:G4}). To create a large dataset with a diverse representation of PIs, only the visual appearance of the PIs is relevant. The underlying experimental conditions and hardware used in the individual acquisition experiments are not critical to this approach. However, a brief overview of the acquisition experiments is provided below:
\begin{enumerate}
    \item \label{exp:G1} Experiment G1 (\textit{Tracer Particles}): Polyamid tracers (particle size $d_\mathrm{P}=20$\,\textmu m, PI size $d_\mathrm{PI}=[19,273]$\,px) in a rotating flow within a glass cylinder. A total of 22,804 PIs, like the ones shown in Fig.\,\ref{fig:Chap5_GAN-Train_ExpG1}, were sampled. The experimental setup of the experiment is described in Lange\,\textit{et al.}\,\cite{LangeSax2024b_conf}.
    
    \item \label{exp:G2} Experiment G2 (\textit{Tracer Particles}): DEHS tracers (PI size $d_\mathrm{PI}=[15,22]$\,px) in an airflow behind a plasma actuator. A total of 11,649 PIs, like the ones shown in Fig.\,\ref{fig:Chap5_GAN-Train_ExpG2}, were sampled. The images were sourced from the experiment described by Pasch\,\textit{et al.}\,\cite{Pasch2024}.

    \item \label{exp:G3} Experiment G3 (\textit{Bubbles}): Bubble column in a rectangular glass tank (bubble size $d_\mathrm{P}=[20,250]$\,\textmu m, PI size $d_\mathrm{PI}=[110,122]$\,px) . A total of 11,875 PIs, like the ones shown in Fig.\,\ref{fig:Chap5_GAN-Train_ExpG3}, were sampled. The images were sourced from the experiment described in Sax\,\textit{et al.}\,\cite{Sax2023d}.

    \item \label{exp:G4} Experiment G4 (\textit{Bubbles}): The same setup as Exp.\,G\ref{exp:G1} was used, but bubbles (particle size $d_\mathrm{P}=[20,300]$\,\textmu m, PI size $d_\mathrm{PI}=[19,273]$\,px) were recorded instead of tracers. A total of 6,721 PIs, like the ones shown in Fig.\,\ref{fig:Chap5_GAN-Train_ExpG4}, were sampled.
\end{enumerate}
Further details on the experiments can be found in the referenced literature \cite{Pasch2024,Sax2023d,LangeSax2024b_conf}.

The two training datasets (Exp.\,G\ref{exp:G1} and G\ref{exp:G2}) contained a total of 34,453 examples of tracer PIs used to train the tracer GAN, while 18,596 examples of bubble PIs from Exp.\,G\ref{exp:G3} and G\ref{exp:G4} were used to train the dispersed phase GAN.

\subsubsection*{Training of the Generative Adversarial Networks}

\begin{figure}%[htbp]
    \centering
    \begin{minipage}{0.5\linewidth}
        \centering
        \begin{subfigure}[b]{0.48\linewidth}
            \centering
            \includegraphics[width=\linewidth]{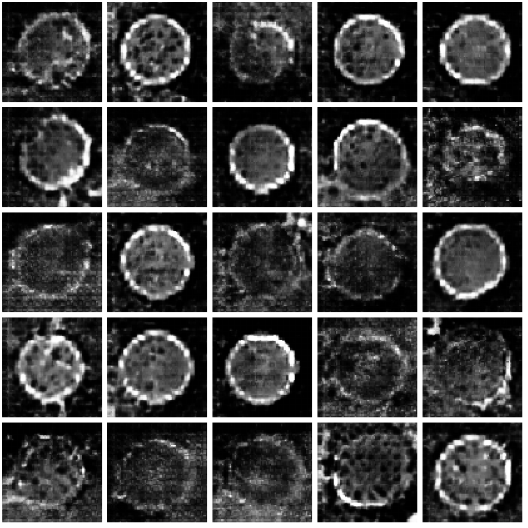}
            \caption{GAN Tracers}
            \label{fig:Chap5_GAN_ExampleIm_DPTV}
        \end{subfigure}
        \hfill
        \begin{subfigure}[b]{0.48\linewidth}
            \centering
            \includegraphics[width=\linewidth]{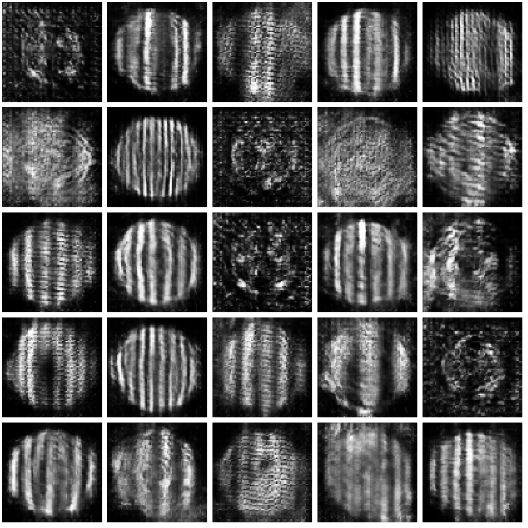}
            \caption{GAN Dispersed Phase}
            \label{fig:Chap5_GAN_ExampleIm_IPI}
        \end{subfigure}
    \end{minipage}
    \caption{GAN-generated images of PIs of tracer particles (a) and particles of the dispersed phase (b). All images have dimensions of $64\times64$\,px.}
    \label{fig:Chap5_GAN_ExampleIm}
\end{figure}

The GANs for tracers and dispersed phase particles share the same architecture and differ only in the training datasets used. MATLAB’s framework and GAN architecture were employed for this purpose \cite{MatlabGAN}. Details of the generator and discriminator architectures are provided in Tab.\,\ref{tab:Appendix_Chap5_GAN_Specifivations} in Appendix\,\ref{Appendix:Chap5_Two-Phase_DPTV}.
Both GANs were trained for 50 epochs. The training specifications are listed in Tab.\,\ref{tab:Appendix_Chap5_GAN_Training}, and the generator and discriminator scores during training are shown in Fig.\,\ref{fig:Appendix_Chap5_GAN_Training_Loss} in Appendix\,\ref{Appendix:Chap5_Two-Phase_DPTV}.
The GAN for the dispersed phase, which was trained on the smaller of the two datasets, exhibited a divergence between the generator and discriminator scores beyond the 29th epoch, as shown in Fig.\,\ref{fig:Appendix_Chap5_GAN_Training_Loss}. This behaviour, indicative of degradation, led to the selection of the checkpoint from the 29th epoch for further use. Notably, the discriminator score approached 0.5 during training, indicating that the discriminator became increasingly uncertain in distinguishing real from generated data—an expected sign that the generator was producing more realistic outputs.
%results 
The resulting PIs generated by both GANs are shown as examples in Fig.\,\ref{fig:Chap5_GAN_ExampleIm_DPTV} for tracers and in Fig.\,\ref{fig:Chap5_GAN_ExampleIm_IPI} for dispersed phase PIs.

\subsection{Training of Object Detection Networks}
\label{chap5:Sec:Approach:Sub:CNNTraining}
%dataset 
Using the outlined approach to generate auto-labeled training data, 20,000 images containing half a million PIs were created. These images were augmented through random vertical and horizontal flipping, resulting in a dataset of approximately 2 million PI examples, equally distributed between the two classes.

\subsubsection*{Selection of Object Detection Networks}
Two different object detection architectures were evaluated to assess the feasibility of two-phase DPTV using CNNs. However, these architectures serve as representative examples, and other models may be used as well.
%Arg 1 single and two step model
Faster R-CNN \cite{Ren.2017} and YOLOv4 \cite{Bochkovskiy2020} were selected to include both a two-stage and a single-stage object detector, respectively.  
%Arg 2: both have been used 
These two networks were chosen, as both have previously been employed for single phase DPTV \cite{Cierpka.2019,Konig.2020,Dreisbach.2022,Sax2023c}. 
%Faster R-CNN choice
The Faster R-CNN network, using a ResNet50 backbone \cite{He2016} (42 million parameters), was trained with the TensorFlow framework \cite{tensorflow2015-whitepaper}. This architecture has demonstrated strong performance in both DPTV and APTV applications \cite{Cierpka.2019,Dreisbach.2022,Sax2023c}.
%YOLO choice 
While YOLOv3 \cite{Redmon2018} has shown inferior results compared to Faster R-CNN in DPTV tasks \cite{Dreisbach.2022}, YOLOv4 \cite{Bochkovskiy2020} incorporates architectural improvements that makes it a promising candidate. YOLOv4 introduces spatial pyramid pooling (SPP) \cite{He2015}, which enhances the network's ability to detect objects at multiple scales, particularly smaller ones. Additionally, YOLOv4 replaces the multi-scale detection approach of YOLOv3 with a more advanced path aggregation network (PANet) \cite{Liu2018}, improving feature fusion across scales. These enhancements result in finer-resolved feature maps with higher semantic information, which were hypothesized to improve PI detection in DPTV \cite{Dreisbach.2022}. YOLOv4 also employs cross stage partial (CSP) connections \cite{Wang2019}, which improves generalization capability.
To investigate whether large backbones are necessary for DPTV, two versions of YOLOv4 were trained. The first used a large CSP-DarkNet-53 backbone, a combination of DarkNet-53 \cite{Redmon2018} and CSP-Net \cite{Wang2019}, with 63.9 million parameters. The second used a significantly smaller CSP-DarkNet-19-tiny backbone with 5.9 million parameters for comparison.

\subsubsection*{Configuration and Training Process}
%training setup
The training was initialized from weights pre-trained on MS-COCO 2018 \cite{Lin.2015} for both networks, to leverage generic features. In this transfer learning approach \cite{Cui.2018}, the networks only need to learn domain-specific features for detecting PIs in each phase of DPTV, which typically results in shorter training times. The training parameters are detailed in Tab.\,\ref{tab:Appendix_Chap5_TrainingMetrics_ObjectDet} in the Appendix\,\ref{Appendix:Chap5_Two-Phase_DPTV}. The training dataset was split (prior to augmentation) into 90\% for training and 10\% withheld for the evaluation of the validation loss. Hyperparameters were selected based on the recommended settings used during pre-training on MS-COCO.

%Faster R-CNN
Faster R-CNN was configured to train for 20,000 iterations without early stopping, based on insights from previous work using this architecture for DPTV \cite{Sax2023c}.
%tiny YOLO 
Both YOLOv4 variants were set up with a maximum of 20,000 training iterations. However, since tiny YOLOv4 is expected to converge faster than the full YOLOv4 network, early stopping with a patience of 50 validation steps was incorporated to prevent overfitting. During training, the validation loss was evaluated every 50 iterations, and training was automatically terminated if no improvement in validation loss was observed for 50 consecutive evaluations.
As a result, training of tiny YOLOv4 was terminated early at 10,850 iterations, with the final model selected based on the checkpoint with the lowest validation loss. The divergence between training and validation loss suggested overfitting. As a cross check, a second version of tiny YOLOv4 was trained from the 10,850-iteration checkpoint to the full 20,000 iterations. This version performed worse on all test datasets, confirming the suspicion of overfitting, and is therefore not further considered.
%YOLOv4
The larger YOLOv4 model was also trained with the same early stopping strategy but completed the full 20,000 iterations without triggering early termination.

The training losses of all three networks are shown in Fig.\,\ref{fig:Appendix_Chap5_Obj_TrainingLoss} in the Appendix\,\ref{Appendix:Chap5_Two-Phase_DPTV}.
For both YOLOv4 and Faster R-CNN, the total loss did not decrease significantly after 15,000 iterations, suggesting that training had converged.

\section{Detection and Separation of Different Particle Image Types}

To evaluate the performance of the object detection networks, images from various experiments were sampled and manually labeled.

\subsection{Test Dataset}

\begin{figure}[htb]
    \centering
    \begin{subfigure}[b]{0.24\linewidth}
        \centering
        \includegraphics[width=\linewidth]{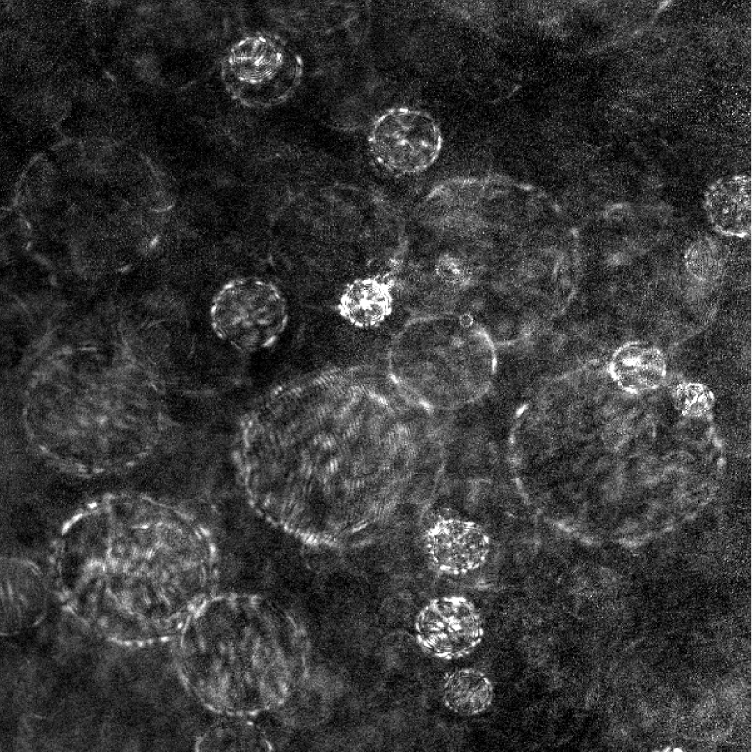}
        \caption{Exp.\,T1 Tracers}
        \label{fig:Chap5_Testimages_T1}
    \end{subfigure}
    \hfill
    \begin{subfigure}[b]{0.24\linewidth}
        \centering
        \includegraphics[width=\linewidth]{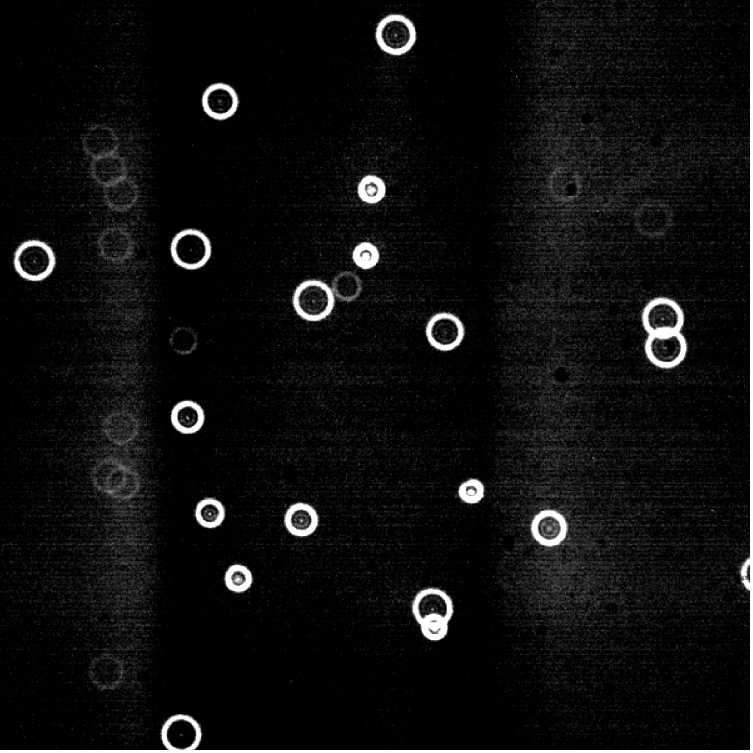}
        \caption{Exp.\,T2 Tracers}
        \label{fig:Chap5_Testimages_T2}
    \end{subfigure}
    \hfill
    \begin{subfigure}[b]{0.24\linewidth}
        \centering
        \includegraphics[width=\linewidth]{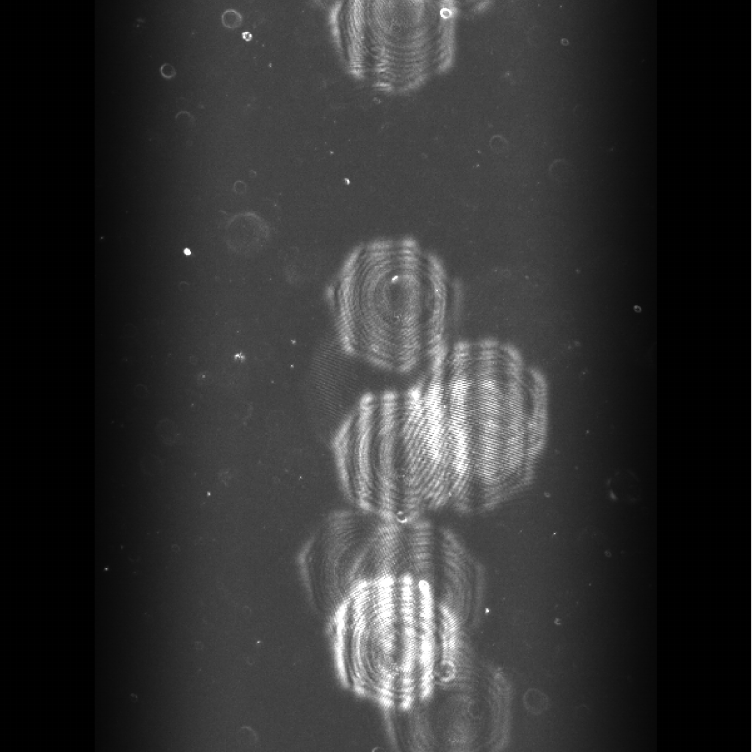}
        \caption{Exp.\,T3 Bubbles}
        \label{fig:Chap5_Testimages_T3}
    \end{subfigure}
    \hfill
    \begin{subfigure}[b]{0.24\linewidth}
        \centering
        \includegraphics[width=\linewidth]{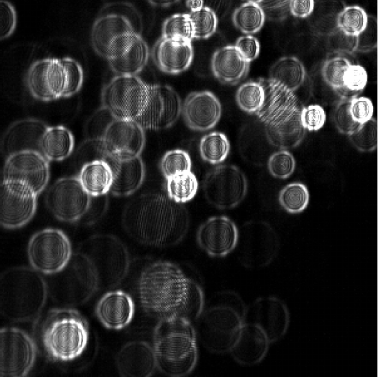}
        \caption{Exp.\,T4 Droplets}
        \label{fig:Chap5_Testimages_T4}
    \end{subfigure}

    \vspace{0.5cm}
    \hspace*{\fill}
    \begin{subfigure}[b]{0.24\linewidth}
        \centering
        \includegraphics[width=\linewidth]{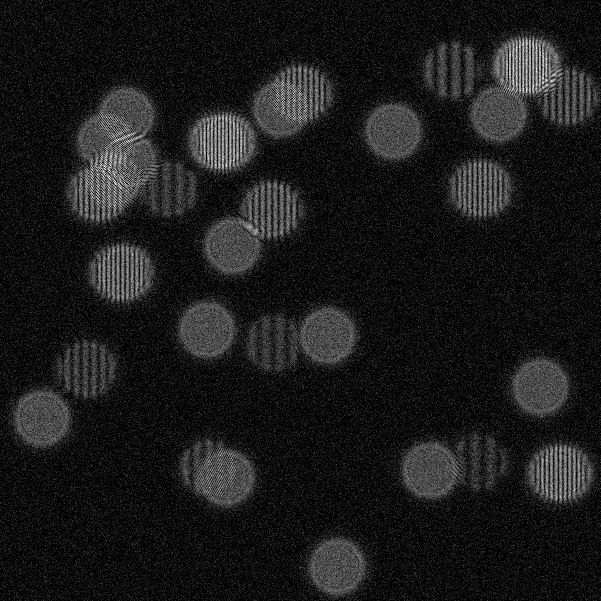}
        \caption{Exp.\,T5 two phase}
        \label{fig:Chap5_Testimages_T5}
    \end{subfigure}
    \hfill
    \begin{subfigure}[b]{0.24\linewidth}
        \centering
        \includegraphics[width=\linewidth]{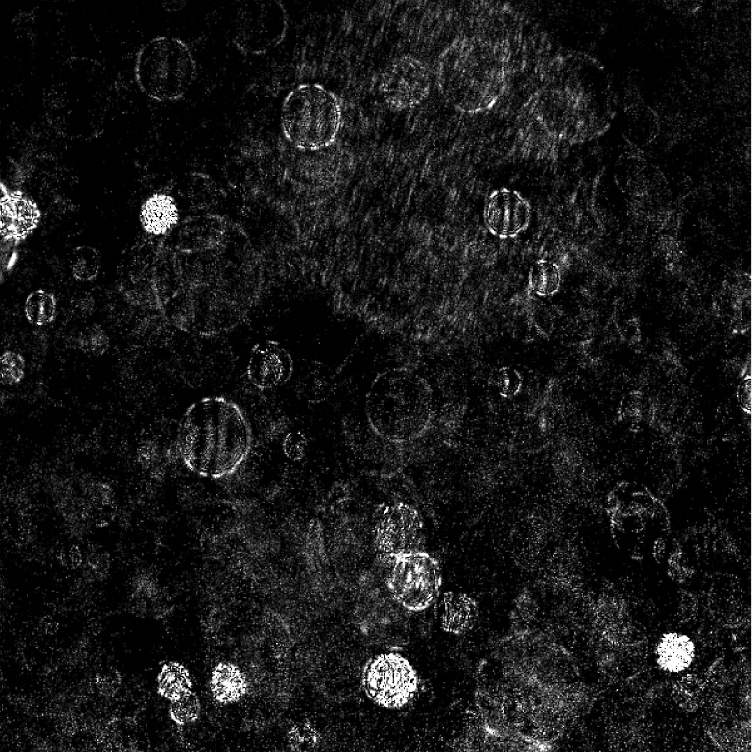}
        \caption{Exp.\,T6 two phase real}
        \label{fig:Chap5_Testimages_T6}
    \end{subfigure} 
    \hspace*{\fill}
    %\vspace{0.5cm}
    \caption{Exemplary images from the six test datasets: T1\cite{LangeSax2024b_conf} (a) and T2\cite{Leister.2021} (b) show tracer particles in a single phase flow. T3\cite{Sax2023d} (c) and T4\cite{poeppe2023} (d) show only bubbles and droplets but no tracers. T5 (e) shows a synthetic image of a two phase flow of bubbles and tracers and T6\cite{LangeSax2024b_conf} (f) show a real image of bubbles and tracers.}
    \label{fig:Chap5_Testimages}
\end{figure}

%dataset structure 
Similar to the training dataset, two datasets containing only tracers and two datasets containing only bubbles or droplets were used. For tracers, the test dataset T\ref{exp:T1} (see Fig.\,\ref{fig:Chap5_Testimages_T1}) contains tracer PIs similar to those in G\ref{exp:G1}, while the image set T\ref{exp:T2} (see Fig.\,\ref{fig:Chap5_Testimages_T2}) differs significantly from the training data.
For dispersed phase PIs, the test dataset T\ref{exp:T3} (see Fig.\,\ref{fig:Chap5_Testimages_T3}) contains bubble PIs similar to those in G\ref{exp:G3}, while the image set T\ref{exp:T4} (see Fig.\,\ref{fig:Chap5_Testimages_T4}) contains droplet PIs that differ significantly from the training data. This was done to test on both familiar and unfamiliar data.

%why single phase
The advantage of using images containing only a single phase is that the ground truth class of each PI is known with absolute certainty. This would not be the case in real images with two types of PIs, where human error in manual labeling could distort the evaluation.
Since the object detector was not informed that only one type of PI was present in each image, the classification performance remains valid. The detector processes each instance independently, without relying on the overall image composition, so the absence of a second class does not affect the validity of the results.

The object detectors were also tested on synthetic (Exp.\,T\ref{exp:T5}, see Fig.\,\ref{fig:Chap5_Testimages_T5}) and real experimental images (Exp.\,T\ref{exp:T6}, see Fig.\,\ref{fig:Chap5_Testimages_T6}) containing two types of PIs. 
For the synthetic images, the ground truth class labels are known, allowing classification performance to be evaluated. In contrast, in the real mixed-phase dataset (T\ref{exp:T6}), BBs could be drawn; however, the manual class labeling of PIs proved unreliable. To avoid drawing conclusions based on potentially flawed ground truth, only the detection but not the classification performance was assessed for this dataset.

Images for the test dataset were drawn from the following experiments and synthetic data generation:
\begin{enumerate}
    \item \label{exp:T1} Experiment T1 (\textit{Tracer Particles}): This experiment used the same setup as Exp.\,G\ref{exp:G1}, but the images were withheld from GAN training. A total of 773 PIs was manually labeled. An example image is shown in Fig.\,\ref{fig:Chap5_Testimages_T1}. The experimental setup of the experiment is described in Lange\,\textit{et al.}\,\cite{LangeSax2024b_conf}.
    
    \item \label{exp:T2} Experiment T2 (\textit{Tracer Particles}): Tracers (particle size $d_\mathrm{P}=9.84$\,\textmu m, PI size $d_\mathrm{PI}=[15,31]$\,px) in an open wet clutch. More details can be found in Leister\,\textit{et al.}\,\cite{Leister.2021}. A total of 394 PIs was manually labeled. An example image is shown in Fig.\,\ref{fig:Chap5_Testimages_T2}.

    \item \label{exp:T3} Experiment T3 (\textit{Bubbles}): This experiment used the same setup as Exp.\,G\ref{exp:G3}, but the images were withheld from GAN training. More details can be found in Sax\,\textit{et al.}\,\cite{Sax2023d}. A total of 202 PIs was manually labeled. An example image is shown in Fig.\,\ref{fig:Chap5_Testimages_T3}.

    \item \label{exp:T4} Experiment T4 (\textit{Droplets}): Water droplets (particle size $d_\mathrm{P}=[20,60]$\,\textmu m, PI size $d_\mathrm{PI}=[16,30]$\,px)  in a spray downstream of a nozzle. The images were sourced from the experiment described in Pöppe\,\cite{poeppe2023}. A total of 554 PIs was manually labeled. An example image is shown in Fig.\,\ref{fig:Chap5_Testimages_T4}.  

    \item \label{exp:T5} Experiment T5 (\textit{Two-phase - synthetic}): Synthetic image of bubbles (simulated particle size $d_\mathrm{P}=[20,250]$\,\textmu m, PI size $d_\mathrm{PI}=[43,112]$\,px) and tracers (PI size $d_\mathrm{PI}=[43,112]$\,px). The forward model described in Sax\,\textit{et al.}\,\cite{Sax2025_IP} was used with the experimental settings from Exp.\,G\ref{exp:G3} (see \cite{Sax2023d}) to generate the images. A total of 3,000 PIs was labeled. An example image is shown in Fig.\,\ref{fig:Chap5_Testimages_T5}.

    \item \label{exp:T6} Experiment T6 (\textit{Two-phase - real}): Tracers and bubbles in the same experimental setups as Exp.\,G\ref{exp:G1} and G\ref{exp:G4} (same tracer and bubble sizes, same PI sizes). A total of 304 PIs was manually labeled. An example image is shown in Fig.\,\ref{fig:Chap5_Testimages_T6}. No class labels were taken from this dataset.
\end{enumerate}
%More details on Exp.\,T\ref{exp:T1}, T\ref{exp:T5}, and T\ref{exp:T6} can be found in Appendix~\ref{Appendix:Chap5_Two-Phase_DPTV}.

The classification performance was rigorously evaluated on synthetic and real single-phase datasets, where class labels are known with absolute certainty. The synthetic dataset consists of idealized PIs that differ only in the presence or absence of fringe patterns, making it particularly well-suited to assess whether the networks have learned this key distinguishing feature. Importantly, the object detector was also tested on the real mixed-phase dataset to evaluate detection performance, ensuring that its ability to generalize to realistic conditions was assessed.

\subsection{Evaluation Metrics}

The evaluation of the neural networks for two-phase DPTV is divided into two parts, assessing detection and classification performance separately. This separation enables a clearer understanding of the network’s ability to distinguish between phases in two-phase DPTV, independent from the influence of PI detection.

%detection
The detection performance is evaluated in a class-agnostic manner. A detection is counted as a true positive ($\mathrm{TP}_\mathrm{D}$) if the predicted BB has an IoU greater than 0.5 with a ground truth (GT) BB, regardless of the predicted class (class-agnostic). This IoU threshold is chosen for comparability with previous work \cite{Dreisbach.2022,Sax2023c}.
A class-agnostic confusion matrix is computed: detections without a corresponding GT BB are counted as false positives ($\mathrm{FP}_\mathrm{D}$), and missed GT BBs are counted as false negatives ($\mathrm{FN}_\mathrm{D}$). Detection performance is typically evaluated using precision $P$
\begin{equation}
    \mathrm{Pr}=\frac{\mathrm{TP}_\mathrm{D}}{\mathrm{TP}_\mathrm{D}+\mathrm{FP}_\mathrm{D}},
    \label{Equ:Precision_Def}
\end{equation}
which represents the probability that a prediction is correct, and recall $\mathcal{R}$
\begin{equation}
    \mathcal{R}=\frac{\mathrm{TP}_\mathrm{D}}{\mathrm{TP}_\mathrm{D}+\mathrm{FN}_\mathrm{D}}, 
    \label{Equ:Recall_Def}
\end{equation}
which is the fraction of detected PIs. However, both precision and recall depend on the chosen confidence score threshold, which reflects the network’s confidence in a given prediction. To eliminate this dependency, the average precision (AP) is computed from precision and recall values across confidence thresholds ranging from 0.001 to 0.999, using the trapezoidal rule:
\begin{equation}
    \text{AP} = \int_0^1 \mathrm{Pr}(\mathcal{R}) \, \mathrm{d}\mathcal{R}
\end{equation}
The class-agnostic AP provides a comprehensive, threshold-independent metric for evaluating detection performance. 
However, AP penalizes the network if it fails to achieve high recall, even when precision is high. Since a detector can still be reliable within lower recall ranges and some missed detections may be acceptable, the truncated AP (TAP) is also computed and normalized by the max recall:
\begin{equation}
    \text{norm TAP} = \frac{1}{\max(\mathcal{R})}\int_0^{\max(\mathcal{R})} \mathrm{Pr}(\mathcal{R}) \, \mathrm{d}\mathcal{R},
\end{equation}
which measures class-agnostic detection performance up to the maximum achieved recall. As TAP depends on the maximum recall reached, both TAP and the maximum recall are considered together.

%classification
To investigate the classification performance independently from detection, a class confusion matrix is computed based on all $\mathrm{TP}_\mathrm{D}$ detections. No confusion matrix is computed for FP detections, as there is no GT class; for FN detections, since no class was predicted; or for true negatives (TN), as neither GT class nor predictions exist.

The detector’s ability to classify a detection correctly is then evaluated using the classification accuracy
\begin{equation}
    \mathrm{CA} = \frac{\mathrm{TP}_\mathrm{C}+\mathrm{TN}_\mathrm{C}}{\mathrm{total\,\, number\,\,of\,\,}\mathrm{TP}_\mathrm{D}}.
\end{equation}
This represents the fraction of correctly classified detections among all TP detections and can be interpreted as the probability that a TP detection is also classified correctly by the detector.
Since the classification accuracy also depends on the confidence score threshold, it is evaluated at the threshold corresponding to the maximum F1-score
\begin{equation}
    \text{F1-score} = 2 \,\frac{\mathrm{Pr}\times\mathcal{R}}{\mathrm{Pr}+\mathcal{R}}
\end{equation}
which identifies the point at which precision and recall are optimally balanced.
The classification accuracy only provides insight into the classification of TP detections, but a detector may also produce FP detections. Since there is no GT class to evaluate the classification of FP predictions, the class bias is computed for $\mathrm{FP}_\mathrm{D}$ predictions instead of the accuracy via
\begin{equation}
    \mathrm{Class\,Bias} = \frac{\mathrm{FP}_\mathrm{D,Class1}-\mathrm{FP}_\mathrm{D,Class2}}{\mathrm{FP}_\mathrm{D,total}}.
\end{equation}
The class bias takes a value of +1 if all FP detections are predicted as tracers, and -1 if all are predicted as dispersed phase, and is zero if the FPs are equally balanced between classes. This provides insight into whether the detector is biased toward predicting one class over the other. Class bias is also evaluated at the threshold corresponding to the maximum F1-score.

Since the localization accuracy of neural network detectors for DPTV and APTV has been extensively studied in previous works (see, e.g., \cite{Cierpka.2019,Konig.2020,Barnkob.2021,Dreisbach.2022,Sax2023c,Ratz2023}), and the primary focus of this study is the distinction between tracers and bubbles or droplets, no further evaluation of localization performance is conducted in this work.

\subsection{Detection Results}

%Detection performance
The results of the networks detection performance is given in Tab.\,\ref{tab:Chap5_NN_DetectionResults}. The networks achieve high average precision across all test sets. Particularly strong performance is observed on images that closely resemble the training data (Exp.,T\ref{exp:T1} and T\ref{exp:T3}). As expected, average precision is lower on the test sets (T\ref{exp:T2} and T\ref{exp:T4}) that diverge more substantially from the training images. This highlights the effect of domain shift between familiar and unfamiliar types of images on the CNN detection performance, and emphasizes the need for the networks to generalize more effectively.
When generalizing to unfamiliar images, the primary limiting factor is the truncation of the maximum achievable recall, while precision remains high. This becomes evident from the norm TAP values, which are close to 1 in most test cases, indicating reliable detection performance within the recall limitations of the network, even on images requiring generalization. This high precision in images visually different from the training data, is an important insight, as high precision helps prevent false positives and, consequently, ghost particles in the tracking step, which could lead to erroneous velocity vectors. Preventing such ghost particles (i.e., achieving high precision) is typically more critical than achieving high recall.

\begin{table}[htb]
    \centering
    \caption{Class agnostic PI detection performance for all six test sets. Shown are the average precision, the max recall and the normalized truncated average precision. The best score for each metric and each test set is highlighted in bold print.}
    \label{tab:Chap5_NN_DetectionResults}
    \begin{tabular}{c|c|c|c|c}
        Test Set & Metric & Faster R-CNN & tiny YOLOv4 & YOLOv4 \\\toprule
        Exp.\,T\ref{exp:T1} & Average precision & \textbf{0.885} & 0.864 & 0.856 \\
        \textit{Tracers only} & max Recall & 0.885 & 0.864 & \textbf{0.892} \\
         & norm TAP & \textbf{1} & 1 & 0.960 \\\midrule
        
        Exp.\,T\ref{exp:T2}  & Average precision & 0.667 & 0.713 & \textbf{0.729} \\
        \textit{Tracers only} & max Recall &0.667 & 0.714 & \textbf{0.729} \\
        & norm TAP & \textbf{1} & 0.984 & \textbf{1} \\\midrule

        Exp.\,T\ref{exp:T3} & Average precision & 0.952 & \textbf{0.989} & 0.957 \\
        \textit{Bubbles only} & max Recall & 0.952 & \textbf{0.989} & 0.957 \\
         & norm TAP & \textbf{1}& \textbf{1} & \textbf{1} \\\midrule
        
        Exp.\,T\ref{exp:T4} & Average precision & 0.765 & 0.819 & \textbf{0.850} \\
         \textit{Droplets only} & max Recall & \textbf{0.862} & 0.819 & 0.850\\
        & norm TAP & 0.888 & \textbf{1} & \textbf{1}  \\\midrule

        Exp.\,T\ref{exp:T5} & Average precision & \textbf{0.997} & 0.993 & 0.993\\
        \textit{Tracers \& Bubbles Syn}& max Recall & \textbf{0.997} & 0.993 & 0.993 \ \\
        & norm TAP & \textbf{1} & \textbf{1} & \textbf{1} \\\midrule
        
        Exp.\,T\ref{exp:T6}  & Average precision & \textbf{0.891} & 0.736 & 0.826 \\ 
        \textit{Tracers \& Bubbles Real} & max Recall & 0.891 & 0.900 & \textbf{0.916} \\
        & norm TAP & \textbf{1} & 0.818 & 0.903 \\\bottomrule
    \end{tabular}
\end{table}

%Classification performance
Following the detection, the next important step is the phase separation. The results for the PI classification are given in Tab.\,\ref{tab:Chap5_NN_ClassificationResults}. Generally, high classification accuracy is observed, with values typically ranging between 95-100\%. The exception is the performance of Faster R-CNN on Exp.\,T\ref{exp:T4} and YOLOv4 on Exps.\,T\ref{exp:T4} and T\ref{exp:T5}, both of which are test sets that require greater generalization.
Tiny YOLOv4, however, showed consistent classification performance across all test images, regardless of the similarity of their visual appearance to the training data.

%Generalization to synthetic Data
Exp.\,T\ref{exp:T5} provides a particularly interesting case, as the images are nearly noise-free and the PIs are represented with idealized features, free from aberrations or other effects that the network might otherwise use as cues. At the same time, the visually distinct appearance of synthetic images to the training data, requires the network to generalize. The strong classification performance (class accuracy of 99.2\% for Faster R-CNN and 95.5\% for tiny YOLOv4) indicates that the networks have learned physically meaningful features, specifically, the ability to distinguish PIs from different phases based on the presence of fringes in the PI.

\begin{table}[htb]
    \centering
    \caption{Classification results on the six different test datasets. The class accuracy for true positive detections and the class bias of false positives is given. The bias becomes +1 if all FP are predicted to be tracers and -1 if predicted as dispersed phase. The best score for each metric and each test set are highlighted in bold print.}
    \label{tab:Chap5_NN_ClassificationResults}
    \begin{tabular}{c|c|c|c|c}
        Test Set & Metric & Faster R-CNN & tiny YOLOv4 & YOLOv4 \\\toprule
        Exp.\,T\ref{exp:T1} & Class Accuracy (TP) & \textbf{0.971} & 0.961 & 0.970 \\
        \textit{Tracers only} & FP Class Bias & +0.653 & +0.868 & \textbf{+0.639} \\\midrule
        
        Exp.\,T\ref{exp:T2}  & Class Accuracy (TP) & 0.983 & 0.998 & \textbf{1}\\
        \textit{Tracers only} & FP Class Bias & \textbf{+0.365} & +0.994 &  +0.915 \\\midrule

        Exp.\,T\ref{exp:T3} & Class Accuracy (TP) & \textbf{1} & 0.994 & \textbf{1} \\
        \textit{Bubbles only} & FP Class Bias & -0.640 & \textbf{-0.468} & -0.658 \\\midrule
        
        Exp.\,T\ref{exp:T4} & Class Accuracy (TP) & 0.754 & \textbf{0.966} & 0.814 \\
         \textit{Droplets only} & FP Class Bias & \textbf{-0.136} & -0.509 & -0.163 \\\midrule

        Exp.\,T\ref{exp:T5} & Class Accuracy (TP) & \textbf{0.992} & 0.955 & 0.824 \\
        \textit{Tracers \& Bubbles Syn} & FP Class Bias & -0.622 & -0.307 & \textbf{-0.264} \\\midrule
        
        Exp.\,T\ref{exp:T6}  & Class Accuracy (TP) & - & - & - \\ 
        \textit{Tracers \& Bubbles Real} & FP Class Bias &-0.357 & \textbf{-0.229} & -0.410\\\bottomrule
    \end{tabular}
\end{table}

%FP Class Bias 
The previous results suggest a low number of false positive (FP) detections, indicated by the norm TAP values close to 1, at reasonable recall levels and reliable phase separation for true positive (TP) detections. However, the networks still exhibit class bias when making FP detections.  
It is important to investigate FP class bias, as false positives introduce non-existent particles into the tracking process, which can influence the accuracy of the determined phase-specific velocity vectors. 
In the single-phase images (Exp.\,T\ref{exp:T1} to T\ref{exp:T4}), the FP class bias tends toward the class present in the image. This is primarily a result of duplicate detections within PI clusters in images where the class distribution is imbalanced.
In contrast, Exps.\,T\ref{exp:T5} and T\ref{exp:T6} contain both classes in balanced proportions. In these mixed-case scenarios, the network tends to show an FP bias toward the dispersed phase, which increases with the confidence threshold (see Fig.\,\ref{fig:Appendix_Chap5_FP_Class_Bias} in Appendix\,\ref{Appendix:Chap5_Two-Phase_DPTV}). Since the training data was class-balanced, a likely explanation is that PIs of bubbles and droplets are more visually distinct than PIs from tracers, leading the network to be more confident in predicting this class.

%Generalization behaviour
The generalization capability to visually different data can be assessed based on the results of Exps.\,T\ref{exp:T2}, T\ref{exp:T4}, T\ref{exp:T5} and T\ref{exp:T6}. It shows that Faster R-CNN generalized better to the synthetic data, indicating that it learned meaningful physical features, whereas YOLOv4 generalized more effectively to different real datasets, both in detection and classification.

\FloatBarrier
\section{Concluding Remarks}

%recap what did we do
%asnwer the main question: does the separation work?
This work demonstrates that pattern-based phase distinction using CNNs is a reliable and effective method for dispersed two-phase DPTV. By leveraging the distinct visual appearance of defocused PIs, the approach enables accurate classification of tracer particles versus bubbles or droplets, achieving 95-100\% accuracy across diverse datasets. When combined with double- or multi-frame setups, this method allows for simultaneous and distinct 3D-3C tracking of both phases using only a single camera. Furthermore, the approach may be extended by incorporating IPI to enable sizing of bubbles or droplets present in the image. 

A key finding is that the CNNs appear to have learned physically meaningful features to distinguish tracers from bubbles or droplets, rather than relying on superficial correlations. A major advantage of the proposed method is that, since the phase distinction is performed entirely in post-processing, it does not require any additional equipment, unlike wavelength-based separation techniques. Furthermore, being a single-camera approach, it enables 3D-3C two-phase tracking with only one optical access, making it particularly attractive for applications with limited optical accessibility.

%advantages 
Beyond its core effectiveness, the method offers several practical advantages. It is particularly well-suited to scenarios where specialized equipment or additional cameras are unavailable or impractical, thereby excluding the use of wavelength-based phase distinction. Unlike size-based separation techniques, the proposed method remains effective even when the size distributions of dispersed and tracer particles overlap. Furthermore, the pattern-based approach does not rely on large velocity differences or high seeding densities, which are typically required for ensemble correlation-based phase separation.
%limitations
However, as a post-processing-based approach, the method depends on the accuracy of the employed CNNs and may be less reliable in phase distinction compared to wavelength-based alternatives. Additionally, the approach entails increased computational demands relative to other post-processing-based phase separation methods.

%autolabeling approach
A central challenge in applying CNN-based phase distinction in two-phase DPTV lies in the availability of sufficiently large and representative training datasets. To address this, the present work introduces a GAN-based auto-labeling framework that enables the generation of realistic, experiment-specific training data from raw, unlabeled images. This approach significantly reduces the manual effort typically required for data annotation and enhances the applicability of CNNs in experimental fluid mechanics.
The method involves conducting single-phase calibration experiments using the same optical setup as the target measurement. The resulting raw images are then used to train a generative model capable of producing large volumes of labeled data that closely resemble the target application. This tailored dataset reduces the domain gap between training and deployment, thereby improving the detection and classification performance of the CNNs.
Moreover, the framework can be extended to incorporate data from multiple experimental setups, facilitating the creation of more diverse training datasets. Such diversity enhances the generalization capability of the trained networks, making them more robust across varying experimental setups and reducing the need for retraining. The results presented in this work demonstrate that both Faster R-CNN and YOLOv4 exhibit improved performance on familiar data.
While the auto-labeling approach introduces additional computational cost, requiring the training of both a generative model and a detection network, it eliminates one of the primary bottlenecks in the adoption of CNNs for DPTV: the need for extensive manual labeling. Furthermore, the method is also applicable to single-phase DPTV, where only one GAN is required, further broadening its utility. 
In summary, the proposed auto-labeling framework constitutes a critical step toward scalable and adaptable CNN-based DPTV.

%further insights and future directions
To improve the approach of pattern based phase distinction in two-phase DPTV, the following considerations should be made:
%recall issue
First, while the CNNs demonstrated very high TAP with, this performance was achieved only within limited maximum recall ranges. Generalization to visually unfamiliar images had little impact on precision but significantly constrained recall. Although high precision is generally more critical in DPTV to avoid ghost particles and the resulting erroneous velocity vectors, higher recall remains desirable. Improved recall would increase the density of velocity vectors and reduce the risk of missing substantial portions of the numbers of particles, which could otherwise lead to incomplete flow characterization. Moreover, achieving recall values close to 100\% for the dispersed phase, in combination with particle sizing via IPI, would enable void fraction estimation. Therefore, future efforts should prioritize increasing recall. This could be achieved either by minimizing the required generalization, through training data that more closely represents the target experiment, or by constructing larger and more diverse datasets to improve the network’s generalization capability. Notably, YOLOv4 exhibited better generalization to real data and may be a promising candidate for future development.
% FP class bias
Second, attention should be given to mitigating the FP class bias observed in the CNNs, which tended to favor bubbles or droplets. Although the high precision achieved by the networks resulted in very few FPs, this bias could become significant when operating at higher recall levels. The presence of this bias, despite a class-balanced training dataset, suggests that additional measures are necessary. One potential strategy is to introduce a slight class imbalance in favor of tracer particles during training, thereby increasing the network’s exposure to the class it is less confident in distinguishing.
%build in synthetic images in training 
Finally, incorporating synthetic examples of PIs alongside real ones in the training data may prove beneficial. Exposure to idealized features during training could help the networks learn more physically meaningful representations and reduce reliance on context-based cues.
These directions collectively aim to improve the robustness, generalization, and quantitative reliability of CNN-based phase distinction in two-phase DPTV.

%repeat main takehome 
In summary, this work confirms that pattern-based phase distinction using CNNs is a viable and practical solution for two-phase DPTV, with promising avenues for further improvement through targeted training data and network design.

%plea to sharing images for more diverse images 
%Additionally, raw image data shared by other researchers could be used to build large, unified datasets containing a wide variety of PIs, similar to how MS-COCO \cite{Lin.2015} is used in computer vision. Such community-wide datasets would enable the training of networks with improved generalization capabilities.

\section*{Acknowledgments}

This work was supported by the Deutsche Forschungsgemeinschaft (DFG, German Research Foundation) via Project Grant KR4775/4-1 within the Research Unit FOR 5595 Archimedes (Oil-refrigerant multiphase flows in gaps with moving boundaries — Novel microscopic and macroscopic approaches for experiment, modeling, and simulation) — Project Number 510921053.
The authors furthermore thank Leister \textit{et al.},\cite{Leister.2021},  Pasch \textit{et al.},\cite{Pasch2024} and Pöppe\cite{poeppe2023} for kindly providing experimental raw images as used in this work.

\section*{Author contributions statement}

\textbf{CS:}
Conceptualization, 
Methodology,
Investigation, 
Software, 
Visualization, 
Data Curation, 
Formal Analysis, 
Validation,  
Writing – Original Draft Preparation, Review \& Editing; 
\textbf{JK:}
Conceptualization, 
Methodology,
Formal Analysis, 
Validation, 
Writing – Review \& Editing, 
Funding Acquisition, 
Supervision, 
Project Administration

\bibliography{main}

\newpage
\section{Appendix}
\label{Appendix:Chap5_Two-Phase_DPTV}

\subsection*{GAN Training Configuration and Training Scores}

\begin{figure}[htb]
\centering
    \begin{subfigure}[t]{0.45\textwidth}
        \centering
        \includegraphics[width=\linewidth]{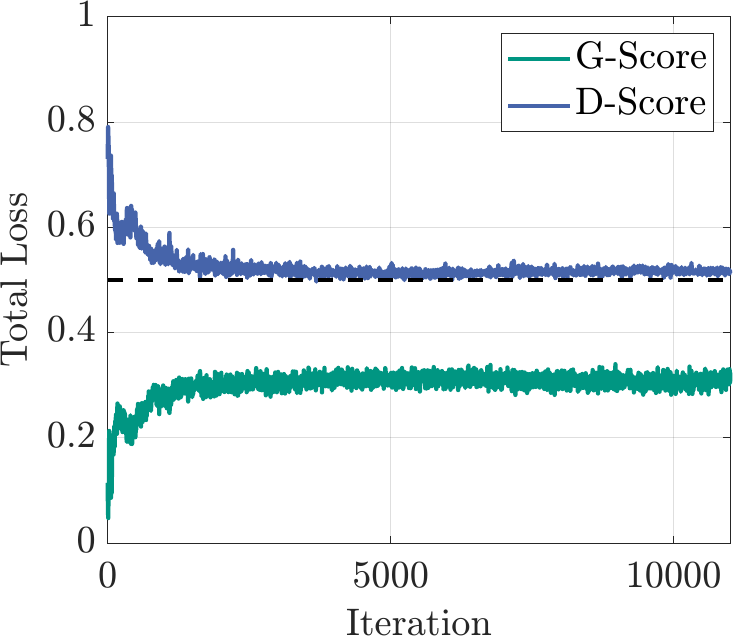}
        \caption{GAN Tracers}
        \label{fig:Appendix_Chap5_GAN_Training_Loss-DPTV}
    \end{subfigure}
    \hfill
    \begin{subfigure}[t]{0.45\textwidth}
        \centering
        \includegraphics[width=\linewidth]{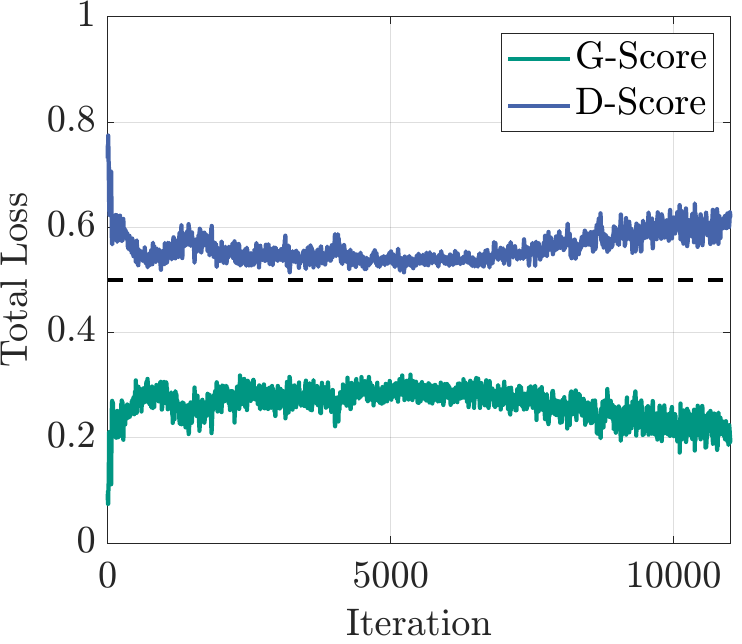}
        \caption{GAN Tracers}
        \label{fig:Appendix_Chap5_GAN_Training_Loss-IPI}
    \end{subfigure}
\caption{Generator score (G-score) and discriminator score (D-score) during the training of the two GANs. Two GANs were trained: One for the generation of PIs representing tracers (a) and another to represent particles of the dispersed phase (b). In both cases the D-Score converged to values close to 0.5 indicating that the generator successfully fools the discriminator.}
\label{fig:Appendix_Chap5_GAN_Training_Loss}
\end{figure}

This section provides further details on the GAN architecture (Tab.~\ref{tab:Appendix_Chap5_GAN_Specifivations}) and the GAN training process (Tab.~\ref{tab:Appendix_Chap5_GAN_Training} and Fig.~\ref{fig:Appendix_Chap5_GAN_Training_Loss}). 
The generator network consists of a project and reshape layer that transforms the latent noise vector, followed by three blocks of transposed convolutional layers, each with batch normalization and ReLU activation. The final layer is a transposed convolution with a \texttt{tanh} activation function.
The discriminator network begins with a dropout layer at the input, followed by four convolutional layers with batch normalization and ReLU activation. The final layer is a convolutional layer with a \texttt{sigmoid} activation function. 
The generator and discriminator architectures used in the present work follow the example provided by MATLAB~\cite{MatlabGAN}.

\begin{table}[htb]
    \centering
    \caption{Specifications of the generator (a) and discriminator (b) network in the GAN, used for both the tracer GAN and the GAN for the dispersed phase. For more information see \cite{MatlabGAN}.}
    \label{tab:Appendix_Chap5_GAN_Specifivations}
    \begin{subtable}[t]{0.45\textwidth}
        \centering
    \begin{tabular}{c|c}
        Network  &  Project and Reshape \\
        Architecture &  3$\times$ (Transposed Conv.\\
                &  Batch Normalization \\
                &      ReLU) \\
                & Transposed Conv. \\
                & tanh \\ 
                & \\ & \\\toprule
        Dim Latent Input & 100 \\
        Output Size & [64,64,1] \\
        Filter Size (Conv) & 5 \\
        Num Filters (Conv) & 64 \\
        Stride (Conv) & 2 \\ & \\ & \\\bottomrule
    \end{tabular}
    \caption{generator}
    \label{tab:Generator_Specifivations}
    \end{subtable}
    \hspace{0.5cm}
    \begin{subtable}[t]{0.45\textwidth}
        \centering
    \begin{tabular}{c|c}
        Network  &  Dropout \\
        Architecture &  Convolution \\
                &  ReLu \\
                & $2\times$ (Convolution \\
                & Batch Normalization \\
                & ReLU) \\
                & Convolution \\
                & Sigmoid \\ \toprule
        Input Size & [64,64,1] \\
        Output & probability \\
        Dropout Probability & 0.5 \\
        nNmFilters (Conv) & 64 \\
        FilterSize (Conv) & 5 \\
        Stride (Conv) & 2 \\
        Scale (Leaky ReLu) & 0.2 \\\bottomrule
    \end{tabular}
    \caption{discriminator}
    \label{tab:Discriminator_Specifivations}
    \end{subtable}
\end{table}

The training parameters for the GANs are provided in Tab.~\ref{tab:Appendix_Chap5_GAN_Training} for both the tracer GAN and the dispersed phase GAN. The GANs were trained using the ADAM optimizer~\cite{Kingma2014}. The generator and discriminator scores during training are shown in Fig.~\ref{fig:Appendix_Chap5_GAN_Training_Loss}.

\begin{table}[htb]
    \centering
    \caption{Training Specifications for the GAN network. For more information see \cite{MatlabGAN}.}
    \label{tab:Appendix_Chap5_GAN_Training}
    \begin{tabular}{c|c|c}
        Network & GAN (Tracer) & GAN (Dispersed Phase) \\ \toprule
        solver & ADAM & ADAM \\
        miniBatchSize & 128 & 128 \\
        learnRate & 2.0000e-04 & 2.0000e-04 \\
        gradientDecayFactor & 0.5000 & 0.5000 \\
        squaredGradientDecayFactor & 0.9990 & 0.9990\\
        labels flipProb & 0.3500 & 0.3500\\
        max Epoch & 50 & 50 \\
        Output Epoch & 50 & 28 \\
        Output Iteration & 11150 & 6756 \\
        data augmentation & horizontal flip & horizontal flip \\
         & vertical flip & vertical flip \\\bottomrule
    \end{tabular}
\end{table}

\FloatBarrier
\subsection*{Training Losses and Configuration for the Object Detection Networks}

\begin{figure} [htb]
\centering
    \begin{subfigure}[t]{0.4\textwidth}
        \centering
        \includegraphics[width=\linewidth]{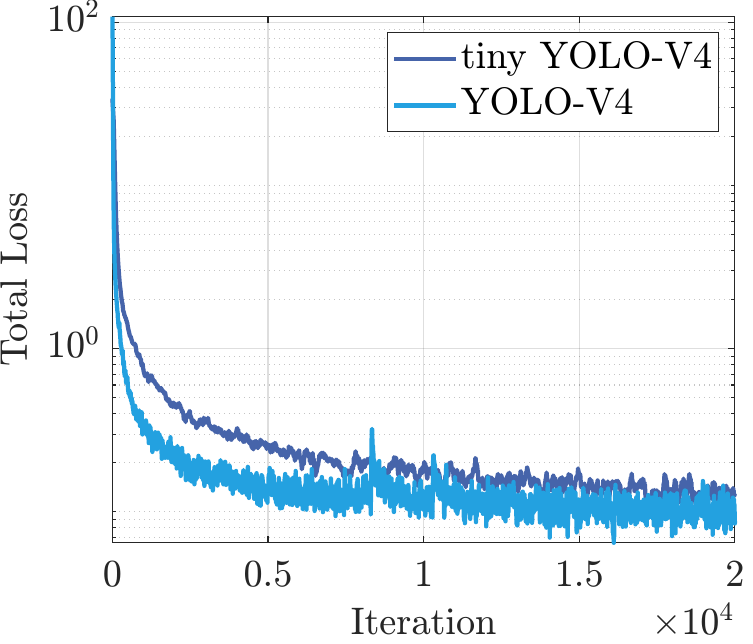}
        \caption{YOLOv4}
        \label{fig:Appendix_Chap5_Obj_TrainingLoss-YOLO}
    \end{subfigure}
    %\hfill
    \begin{subfigure}[t]{0.4\textwidth}
        \centering
        \includegraphics[width=\linewidth]{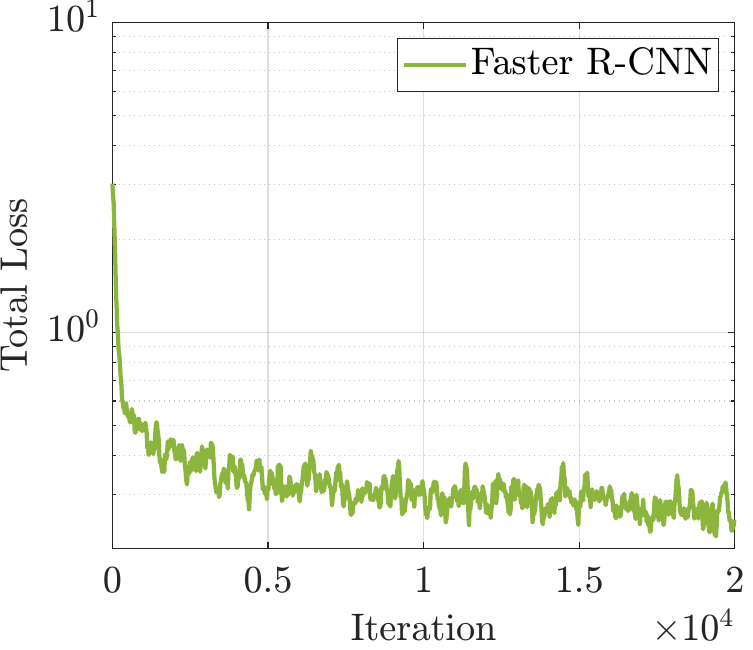}
        \caption{Faster R-CNN}
        \label{fig:Appendix_Chap5_Obj_TrainingLoss-F-RCNN}
    \end{subfigure}
\caption{Total training loss of Faster R-CNN (a) and YOLOv4 and tiny YOLOv4 (b).}
\label{fig:Appendix_Chap5_Obj_TrainingLoss}
\end{figure}

The training configuration and the hyperparameters used to train Faster R-CNN and YOLOv4 are presented in Tab.~\ref{tab:Appendix_Chap5_TrainingMetrics_ObjectDet}. Both networks were trained using stochastic gradient descent with momentum. Pre-trained weights from MS-COCO 2018 were used to initialize the training. The corresponding training losses are shown in Fig.~\ref{fig:Appendix_Chap5_Obj_TrainingLoss}.

\begin{table}[htb]
\begin{center}
\caption{Training settings for Faster R-CNN (ResNet50), Tiny YOLOv4 and YOLOv4 (DarkNet)}
\label{tab:Appendix_Chap5_TrainingMetrics_ObjectDet}
\begin{tabular}{lccc}\toprule
\textbf{$IoU\,\geq\,0.5$} & \textbf{Faster R-CNN} & \textbf{YOLOv4} & \textbf{YOLOv4}\\
	        & ResNet50  & CSPDarkNet53-Tiny &  CSPDarkNet53 \\\midrule
parameters (millions) & 42 & 5.8 & 63.9 \\\midrule
optimizer & SGDM & SGDM & SGDM \\
training iterations & 20,000 & 20,000 & 20,000\\
patience (validation checks) & none & 50 & 50\\
validation frequency & 50 & 50 & 50\\
output iteration & 20,000 & 10,850 & 20,000\\
minibatch size & 4 & 4 & 4 \\
learning rate $\alpha$ & 0.0003 & 0.001 & 0.001\\
learning rate decay & none & none & none \\
momentum $\beta$ & 0.9 & 0.9& 0.9\\
$\mathrm{L_2}$-regularization & none & 0.0005 & 0.0005\\
data augmentation & horizontal flip & horizontal flip & horizontal flip\\
 & vertical flip & vertical flip & vertical flip \\
pre-trained & MS-COCO2018 & MS-COCO2018 & MS-COCO2018 \\\bottomrule
\end{tabular}
\end{center}
\end{table}

\FloatBarrier
\subsection*{False Positive Class Bias}
\FloatBarrier

\begin{figure} [htb]
\centering
    \begin{subfigure}[t]{0.6\textwidth}
        \centering
        \includegraphics[width=\linewidth]{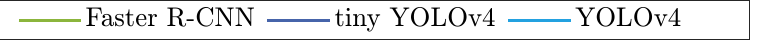}
    \end{subfigure}
    \vskip 0.5cm
    \begin{subfigure}[t]{0.6\textwidth}
        \centering
        \includegraphics[width=\linewidth]{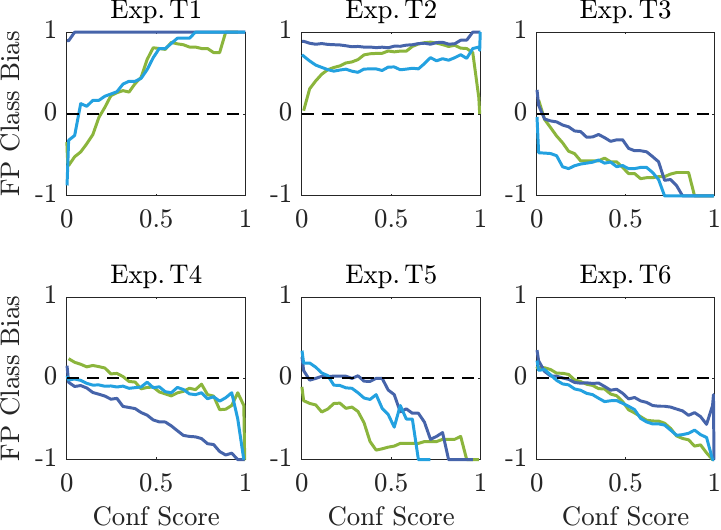}
    \end{subfigure}
\caption{False positive class biases for Exps.\,T1, T2, T3, T4, T5, T6 over the applied confidence score threshold. The black dashed line indicates zero bias. The bias becomes +1 if all FP are predicted to be tracers and -1 if predicted as dispersed phase.}
\label{fig:Appendix_Chap5_FP_Class_Bias}
\end{figure}

For the three object detection networks, Faster R-CNN, YOLOv4, and Tiny YOLOv4, the false positive class bias was computed and is presented in Fig.\,\ref{fig:Appendix_Chap5_FP_Class_Bias}. It can be observed that in Exps.~\ref{exp:T3}, \ref{exp:T4}, \ref{exp:T5}, and \ref{exp:T6}, the class bias starts near zero for confidence scores of 0.0001 and approaches -1 as the confidence score increases to 0.999. This indicates that the networks exhibit increasing confidence in the dispersed phase and are biased toward this class. In the single-phase experiments containing only tracers (Exps.~\ref{exp:T1} and \ref{exp:T2}), a class bias toward the tracer class is observed.

\clearpage

\end{document}